\def\year{2019}\relax
\documentclass[letterpaper]{article} %DO NOT CHANGE THIS
\usepackage{aaai19}  %Required. 
\usepackage{times}  %Required
\usepackage{helvet}  %Required
\usepackage{courier}  %Required
\usepackage{url}  %Required
\usepackage{graphicx}  %Required
\usepackage{subcaption}
\graphicspath{ {./figures/} }
\usepackage{amsmath}
\usepackage{amsfonts}
\usepackage{multirow}
\newcommand{\mE}{\mathbb{E}}

\begin{document}

\title{Guiding the One-to-one Mapping in CycleGAN via Optimal Transport}
%\author{AAAI Submission 5793}
\author{Guansong Lu, Zhiming Zhou, Yuxuan Song, Kan Ren, Yong Yu \\
Shanghai Jiao Tong University \\
\{gslu, heyohai, songyuxuan, kren, yyu\}@apex.sjtu.edu.cn%, \{yxsong0816, rk6556\}@gmail.com
}
\maketitle

\begin{abstract}
CycleGAN is capable of learning a one-to-one mapping between two data distributions without paired examples, achieving the task of unsupervised data translation. However, there is no theoretical guarantee on the property of the learned one-to-one mapping in CycleGAN. In this paper, we experimentally find that, under some circumstances, the one-to-one mapping learned by CycleGAN is just a random one within the large feasible solution space. Based on this observation, we explore to add extra constraints such that the one-to-one mapping is controllable and satisfies more properties related to specific tasks. We propose to solve an optimal transport mapping restrained by a task-specific cost function that reflects the desired properties, and use the barycenters of optimal transport mapping to serve as references for CycleGAN. Our experiments indicate that the proposed algorithm is capable of learning a one-to-one mapping with the desired properties.
\end{abstract}

\section {Introduction}

Image-to-image translation aims at learning a mapping between a source distribution and a target distribution, which can transform an image from the source distribution to that from the target distribution. It covers a variety of computer vision problems including image denoising~\cite{denoising}, segmentation~\cite{segmentation}, and saliency detection~\cite{saliency}. Along with the recent popularity of deep supervised learning, many algorithms based on paired training data and deep convolution neural networks have been proposed for specific image-to-image translation tasks. Among them, Pix2pix~\cite{pix2pix} proposed an image-to-image translation framework utilizing adversarial training technique to force the translation results being indistinguishable from the target distribution. 

In practice, it is usually difficult to collect a large amount of paired training data, while unpaired data can usually be obtained more easily, hence unsupervised learning algorithms have also been widely studied. Particularly, generative adversarial networks (GANs;~\cite{gan}) and dual consistency~\cite{duallearning} are extensively studied in image-to-image translation. CycleGAN~\cite{cyclegan}, DiscoGAN~\cite{disco} and DualGAN~\cite{dualgan} adopt these two techniques for solving unsupervised image-to-image translation where GAN loss is used to ensure the generated images being indistinguishable from real images and cycle consistency loss helps to establish a one-to-one mapping between source distribution and target distribution. In this paper, to simplify the terminology, we will use CycleGAN as a representative for these three similar frameworks combining GANs and the idea of cycle consistency.

CycleGAN can establish a one-to-one mapping between two data distributions unsupervisedly with the help of the cycle consistency losses in both directions. However, theoretically, there is no claim on the detailed properties of the mapping established by CycleGAN, which results in a large feasible solution space. Consequently, without meticulously designed network and hyper-parameters, the one-to-one mapping learned by CycleGAN will be a random one within this large space. 

For many cross-domain translation tasks, people actually have expected properties on the learned mapping, e.g. in language translation task, people would expect the semantic meaning keeps unchanged. Hence, it will be more satisfactory if we can add explicit constraints on the one-to-one mapping within CycleGAN to control the mapping's properties, so as to meet the requirements of specific tasks.

Among the many potential feasible maps between two data distributions, it is more promising to find the optimal one according to some measure. Optimal transport (OT) aims at finding a transportation plan~\cite{kantorovich} that holds the least cost of transporting the source distribution to the target distribution, given a cost function that specifies the transportation cost between any pair of samples from the two distributions. 
%While this transportation plan is usually not a one-to-one mapping, but many-to-many instead, \cite{ferradans2014regularized} proposes to establish barycentric mapping from this transportation plan such that the target for each source is a single point and applies it for color adaptation. \cite{large-scale} also proposes to train a neural network to learn the underlying mapping using the barycentric mapping.

It is worth mentioning that the cost function in optimal transport is very flexible. For specific tasks, it is possible to define a cost function to reflect the underlining expectation of the desired mapping properties. %When applied to two image distributions, the cost function can be specified as any metric between two images according to specific tasks. 
For example, given a set of handbags and shoes, if one would like to pair the handbags with the shoes such that they have matched colors, one can specify the cost function to be the distance between their color histograms, and then the optimal transport would find the mapping that has the least overall difference in color distribution. 
%the mapping to keep the color distribution, i.e. 
%target image and the source image, the mapping, so as to find a mapping that matches the color distribution of the target image and the source image. Strictly speaking, it is the mapping with the least overall difference of color distribution. 

In summary, CycleGAN lacks the control of the one-to-one mapping, while optimal transport holds the ability to establish a mapping towards the desired property. However, the optimal transport mapping, i.e., transportation plan, is usually not a one-to-one mapping, but many-to-many instead; that is, we cannot directly use optimal transport to build a desired one-to-one mapping. We thus propose to use optimal transport as a reference to endow CycleGAN with the ability of learning a one-to-one mapping with desired properties. %guide the learning of the one-to-one mapping in CycleGAN. 

The contributions of this paper has been summarized as below.

\begin{itemize}
\item We study the properties of the one-to-one mapping learned by CycleGAN and verify that under some circumstances the one-to-one mapping learned by CycleGAN is just a random one within the large feasible solution space, which is due to the lack of constraint on the one-to-one mapping established by CycleGAN.
\item We propose to use the optimal transport with respect to a task-specific metric to guide CycleGAN on learning a one-to-one mapping with desired properties. Our experiments on several datasets demonstrate the effectiveness of the proposed algorithm on learning a desired one-to-one mapping.
\end{itemize}

\section {Related Work}

Generative Adversarial Networks (GANs), consisting of a generator network and a discriminator network, is originally proposed as a generative model to match the distribution of generated samples to the real distribution, where the discriminator is trained to distinguish generated samples from real ones while the generator learns to generate samples that fool the discriminator.
% The objective of generator is to generate images whose distribution closely matches the real image distribution while the  discriminator is trained to distinguish generated images from real ones. 
%The generator network is trained by the adversarial loss from the discriminator network to map the random noise distribution to the real image distribution, achieving the task of image generation.
%
Researchers have been working hard on improving the stability of training and exploiting the capacity of GANs for various computer vision tasks. For instance, \cite{dcgan} proposes a deep convolutional architecture that stabilizes the training; WGAN~\cite{wgan} proposes to utilize Wasserstein-1 distance (or Earth Mover's distance/EMD) as an alternative metric. %Compared with original GANs, the training of WGAN is more stable as Wasserstein distance between two data distributions never saturates. % so that the critic in WGAN can provide meaningful gradients to the generator. 

Conditional GANs (cGANs;~\cite{cgan,acgan,amgan}) proposes to extend GANs to a conditional model by conditioning some extra information, such as class label, on both generator and discriminator in GANs so that it can generate images conditioned on class labels and so on. \cite{text2image} extends cGANs with conditional information being text features. Pix2pix~\cite{pix2pix} proposed a unified image-to-image translation framework based on conditional GANs, with conditional information being images.

In practice, it is always hard to collect a large amount of paired training data, while unpaired data can always be obtained more easily. In order to make better use of unpaired data in real world, CycleGAN~\cite{cyclegan}, DiscoGAN~\cite{disco} and DualGAN~\cite{dualgan} adopt the idea of dual consistency, which was firstly proposed in language translation~\cite{duallearning}, together with GANs to simultaneously train a pair of generators and discriminators for translation in both directions and applied cycle consistency loss on both data distributions, which forces the mapping to be a one-to-one mapping. However, theoretically, there is no explicit constraint on the properties of the one-to-one mapping within CycleGAN, which results in a large feasible solution space and the learned one-to-one mapping being a random one within this space.

Optimal transport~\cite{oldandnew} aims to find a mapping between two distributions that can transport the source distribution to the target distribution with the least transportation cost. In many cases, the mapping  between two distributions, where each source point only maps to a single target point (the Monge's problem) does not exist. The modern approach to optimal transport relaxes the Monge's problem by optimizing over plans, i.e., a distribution over the product space of the source distribution space and the target distribution space. %This relaxation casts the optimal transport as a linear program which is always feasible. 
\cite{sinkhorn} proposes to introduce entropic regularization term into OT problem which turns it into an easier optimization problem and can be solved efficiently by Sinkhorn-Knopp algorithm.
%
%The resulting transportation cost and transportation plan are called xxxx distance and optimal transportation plan respectively.
%\cite{ferradans2014regularized} proposes graph regularization term for the task of color transfer and extended this framework to the computation of barycenter of distributions. 
%\cite{joint-coupling-mapping} proposes to jointly learn the transportation plan and the transport map. \cite{courty2017optimal} deals with domain adaptation by solving barycentric mapping based on the optimal transportation plan. 
\cite{large-scale} proposed a stochastic approach for solving large-scale regularized OT and estimating a Monge mapping as a deep neural network approximating the barycentric mapping of the OT plan. %, which also allows the mapping to generalize to samples outside the support of the input measure.

%() proposed to use Wasserstein-1 distance (or Earth Mover’s distance/EMD) between images as a metric for image retrieval. WGAN \cite{wgan} proposed to use Wasserstein-1 distance as a metric between generated data distribution and real data distribution in lieu of the KL divergence or JS divergence in original GANs for stable training.

\section {Method}

\begin{figure*}[t!]
\centering
\includegraphics[width=0.95\textwidth]{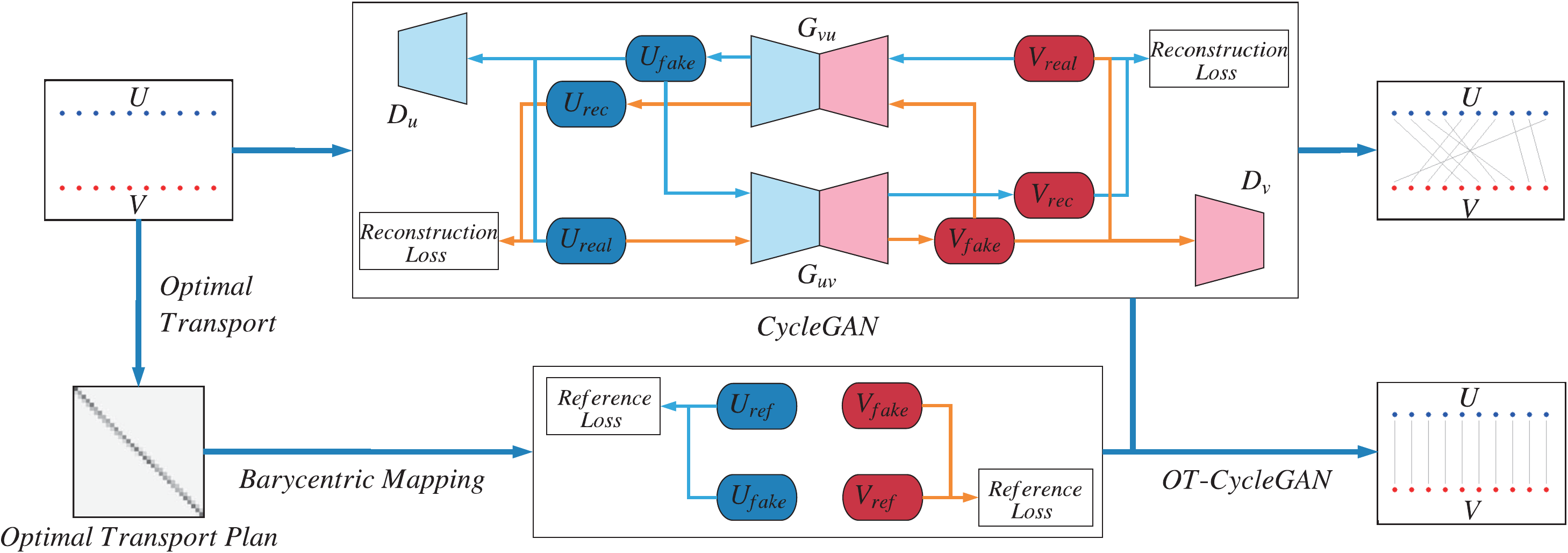}
\caption{The framework of the  proposed method. We use the barycentric mapping of optimal transport, which minimizes the cost of mismatching of a task-specific property, to guide the CycleGAN on learning a one-to-one mapping with the desired properties.}
\label{fig:framework}
\end{figure*}

Given two sets of unpaired images that respectively from domain $U$ and domain $V$, the primal task of unsupervised image-to-image translation is to learn a generator $G_{uv}: U \rightarrow V$ that maps an image $u \in U$ to an image $v \in V$. The modern techniques \cite{cyclegan,dualgan,disco} of unsupervised image-to-image translation introduce an extra generator $G_{vu}: V \rightarrow U$ that maps an image $v \in V$ to an image $u \in U$ and cycle consistency loss, i.e., $G_{vu}(G_{uv}(u))\approx u$ and $G_{uv}(G_{vu}(v))\approx v$, is introduced to regularize the mapping between $U$ and $V$. As the result, the learned mapping would be a bijection, i.e., a one-to-one mapping. However, as we will discuss in the latter of this section, cycle consistency loss, though helps build a one-to-one mapping, has no control on the properties of the learned one-to-one mapping. In this section, we will also discuss how to add extra constraints on the learning of the one-to-one mapping to chase desired properties. %\yuxuan{how to add may not be the key point in this section, add what kind of constraint is the key....I am not sure}

\subsection{Preliminary: CycleGAN}

%Unsupervised image-to-image translation requires the translation result to be vivid. 
In CycleGAN, besides the above-mentioned two coupled generators $G_{uv}$ and $G_{vu}$ that translate images across domain $U$ and $V$ and the cycle consistency losses that regularize the learned mapping to be a bijection, it also introduces an adversarial loss to each generator to ensure translated images are valid samples. More strictly, by playing a minimax game with the discriminator, the adversarial loss forces the generator to match the distribution of generated images with the distribution of real images in the target domain.

\subsubsection{Adversarial Loss}

In the original GAN~\cite{gan}, the discriminator was formulated as a binary classifier outputting a probability. Given a real image distribution $\nu$ and the fake image distribution formed by generated samples $G_{uv}(u)$ with $u \sim \mu$, the loss function of original GAN is defined as:
\begin{equation}
\begin{aligned}
L_{gan}(G_{uv}, D_v)&={\mE}_{u \sim \mu}[\log (1-D_v(G_{uv}(u)))] \\  
                    &+{\mE}_{v \sim \nu}[\log D_v(v)],
\end{aligned}
\end{equation}
% \begin{equation}
% \!\!\!\!\!\!\!\!\!\!\!\!\!\!\!\!\!\!\!\!\!\!\!\!\!\!\!\!\!\!\!\!\!\!\!\!\!\!\!\!\!\!\!\!\!\!\!\!
% L_{gan}(G)=-{\mE}_{x \sim P_{g}}[-log(1-D(x))].
% \end{equation}
The discriminator $D_v$ learns to maximize $L_{gan}(G_{uv}, D_v)$, that is to distinguish the real samples and the fake samples, while the generator learns to minimize $L_{gan}(G_{uv}, D_v)$ such as to make the generated samples have a low probability of being classified as fake by the discriminator. 
%
%We refer to the distribution of $G_{uv}(u)$ with $u \sim \mu$ as $G_{uv}(\mu)$. 
When $D_v$ is assumed to be optimal, the objective of generator is to minimize the Jensen-Shannon divergence between $G_{uv}(\mu)$ and $\nu$, and the minimum is achieved if and only if $G_{uv}(\mu)=\nu$. 

Although GANs have achieved great success in the realistic image generation, training of the original GANs turns out to be very difficult and one has to carefully balance the ability of generator and discriminator. It was showed in \cite{principled_methods,wgan} that Jensen-Shannon divergence is ill-defined when the supports of the two distributions are not overlapped. Wasserstein distance is thus introduced~\cite{wgan} as an alternative metric for evaluating the distance between the real and fake distributions. %, which provides more reliable gradients for the training of generator.
Wasserstein distance $W(\mu,\nu)$ is defined as the minimal cost of transporting distribution $\mu$ into $\nu$. In its primal form, it is formally defined as:
\begin{equation}\label{EqW1}
W(\mu,\nu) = \inf_{\pi \in \Pi(\mu,\nu)} \, {\mE}_{(u,v) \sim \pi} \, [{d(u, v)}],
\end{equation}
where $\Pi(\mu,\nu)$ denotes the collection of all probability measures on $U \times V$  with marginals $\mu$ on $U$ and $\nu$ on V. %Under mild assumptions, $W(\mu,\nu)$ is continuous everywhere and differentiable almost everywhere. 
%denotes the set of all joint distributions with $\mu$ and $\nu$ being the marginal distribution of the first and second factors respectively, \emph{i.e.} $\int_{\mathcal{\nu}} \pi(u, v)dv = \mu(u)$ and $\int_{\mathcal{\mu}} \pi(u, v)du = \nu(v)$. 

Since the infimum in Eq.~(\ref{EqW1}) is highly intractable, in WGAN~\cite{wgan}, the discriminator~(critic) is designed to estimate the Wasserstein distance by solving its dual form, with the corresponding objective defined as: %which is constrained as a 1-Lipschitz function is utilized to evaluate Wasserstein Distance in the dual form, 
\begin{equation}
L_{wgan}(G_{uv}, D_v)={\mE}_{v \sim \nu}[D_v(v)]-{\mE}_{u \sim \mu}[D_v(G_{uv}(u))], \\  %,f \in Lip_{1}
\end{equation}
% \begin{equation}
% \!\!\!\!\!\!\!\!\!\!\!\!\!\!\!\!\!\!\!\!\!\!\!\!\!\!\!\!\!\!\!\!\!\!\!\!\!\!\!\!\!\!\!\!\!\!\!\!
% L_{wgan}(G)=-{\mE}_{x \sim P_{g}}[f(x)].
% \end{equation}
where the discriminator is constrained as a 1-Lipschitz function. The problem of how to properly enforce 1-Lipschitz has evolved a set of investigations \cite{wgangp,sngan,wganlp}. In our experiments, these solutions show very similar results and we choose the Gradient-Penalty \cite{wgangp} loss for on-the-fly example through the paper, i.e.,
\begin{equation}
L_{gp}(D_v)= {\mE}_{\hat{x} \sim P_{\hat{x}}} \, [ (\lVert \nabla D_v(x) \rVert_2 -1)^2],
\end{equation}
where $P_{\hat{x}}$ is the distribution of uniformly distributed linear interpolations of $v \sim \nu$ and $x \sim G_{uv}(\mu)$.

\subsubsection{Cycle Consistency Loss}

%Introduce dual task \cite{duallearning} and adding reconstruction loss \cite{cyclegan,dualgan,disco} has show it power on building cross-domain relationship. %In the context of unsupervised, cross domain data translation, the idea of 

%the idea of dual learning was first proposed in machine translation \cite{duallearning} as a two-agent communication game, in which two agents each knows about only one language and learn to improve the quality of the two translation models collectively. The communication system with the two agent is connected by a machine translation module and formed as a close loop to minimize the loss of information during the translation.

%the adversarial loss itself can not guarantee the one-to-one property of the learned mapping function, to further reduce the space of possible mapping functions,
%The intuition of the reconstruction loss here is similar to dual learning, \emph{i.e.} to retain more information during the mapping.

Training $G_{uv}$ with respect to the adversarial loss forces the distribution of $G_{uv}(u)$ to match with the distribution $\nu$. However, this actually does not build any relationship between the source domain and the target domain. Without paired data, traditional approaches build the relationship between the domain data via predefined similarity function~\cite{AAA,shrivastava2017learning,taigman2016unsupervised} or assuming shared low-dimensional embedding space~\cite{EEE,DDD}. In CycleGAN series~\cite{cyclegan,disco,dualgan}, a dual task of translating data from domain $V$ to domain $U$ is introduced and cycle consistency is encouraged as a regularization. 

Specifically, cycle consistency requires any image $u$ in domain $U$ can be reconstructed after applying $G_{uv}$ and $G_{vu}$ on $u$ in turn and any image $v$ in domain $V$ can be reconstructed after applying $G_{vu}$ and $G_{uv}$ on $v$ reversely. That is, $G_{vu}(G_{uv}(u)) \approx u$, $G_{uv}(G_{vu}(v)) \approx v$. The cycle consistency loss can be formulated as follow:
\begin{equation}
\begin{aligned}
&L_{rec}(G_{uv}) = {\mE}_{u \sim \mu}\left[\| G_{vu}(G_{uv}(u)) - u \|\right], \\
&L_{rec}(G_{vu}) = {\mE}_{v \sim \nu}\left[\| G_{uv}(G_{vu}(v)) - v \|\right],
\end{aligned}
\end{equation}
in which we adopt $L_{2}$ distance to measure the distance between the original image and the reconstructed image.

\subsection{The One-to-one Mapping in CycleGAN}
In CycleGAN, the adversarial losses applied on two generators help to establish the mappings between domain $U$ and domain $V$ in both directions, as it forces the generated images to be within the target domain. Meanwhile, the cycle consistency losses help to relate these two mappings and force them to be two one-to-one mappings, as it forces different samples in the source domain to be mapped to different samples in the target domain (otherwise, the consistency loss would be large). Therefore, CycleGAN would establish a bijective mapping between domain $U$ and domain $V$, which is also mentioned in DiscoGAN \cite{disco} and CycleGAN \cite{cyclegan}.

It is promising that CycleGAN can find a one-to-one mapping between two data distributions unsupervisedly. But theoretically, there exists a large number of one-to-one mappings between two data distributions. For example, the number of possible one-to-one mapping between two discrete data distributions with each containing $n$ discrete data points is the factorial of $n$, i.e. $n!$. And all these one-to-one mappings are perfect in terms of CycleGAN's objective. 

Since there is no extra control on the properties of the mapping, as long as it is one-to-one, the learned one-to-one mapping with CycleGAN would theoretically be a random one in this large feasible solution space.

\begin{figure*}[htbp]
\begin{subfigure}[t]{0.33333\textwidth}
\centering
\raisebox{15pt}{\includegraphics[width=0.99\textwidth]{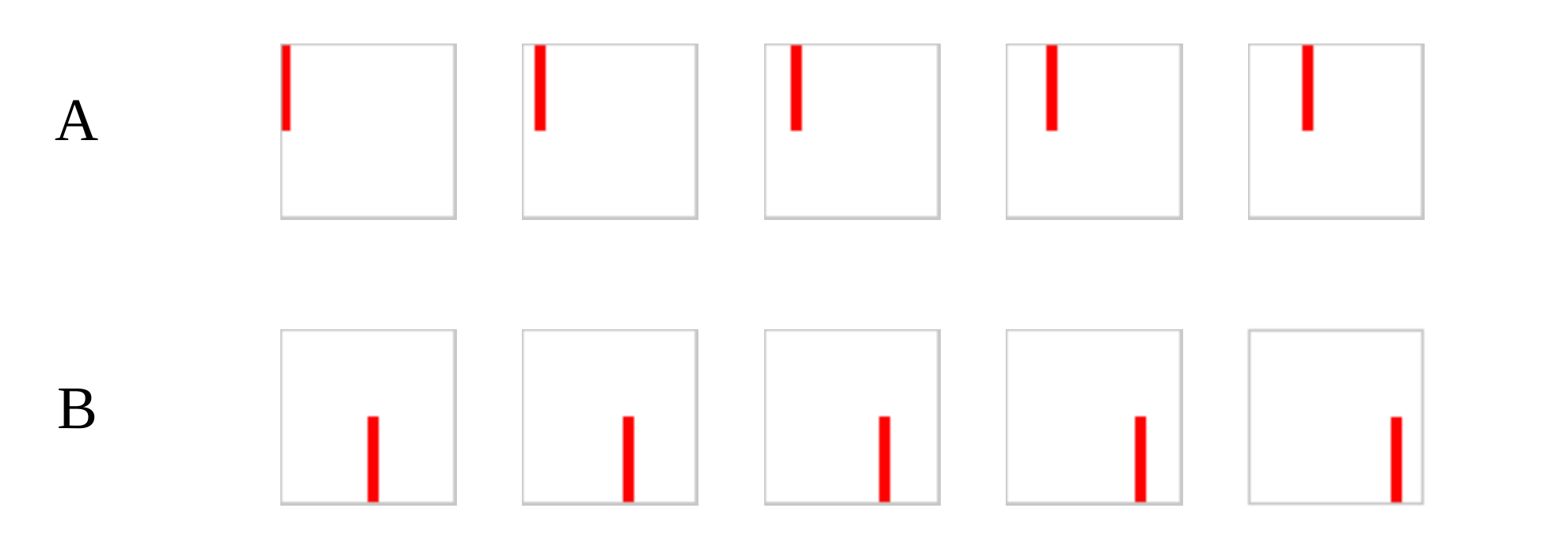}}
\caption{Datasets Samples}
\label{fig:sythetic-a}
\end{subfigure}
\begin{subfigure}[t]{0.33333\textwidth}
\centering
\includegraphics[width=0.5\textwidth]{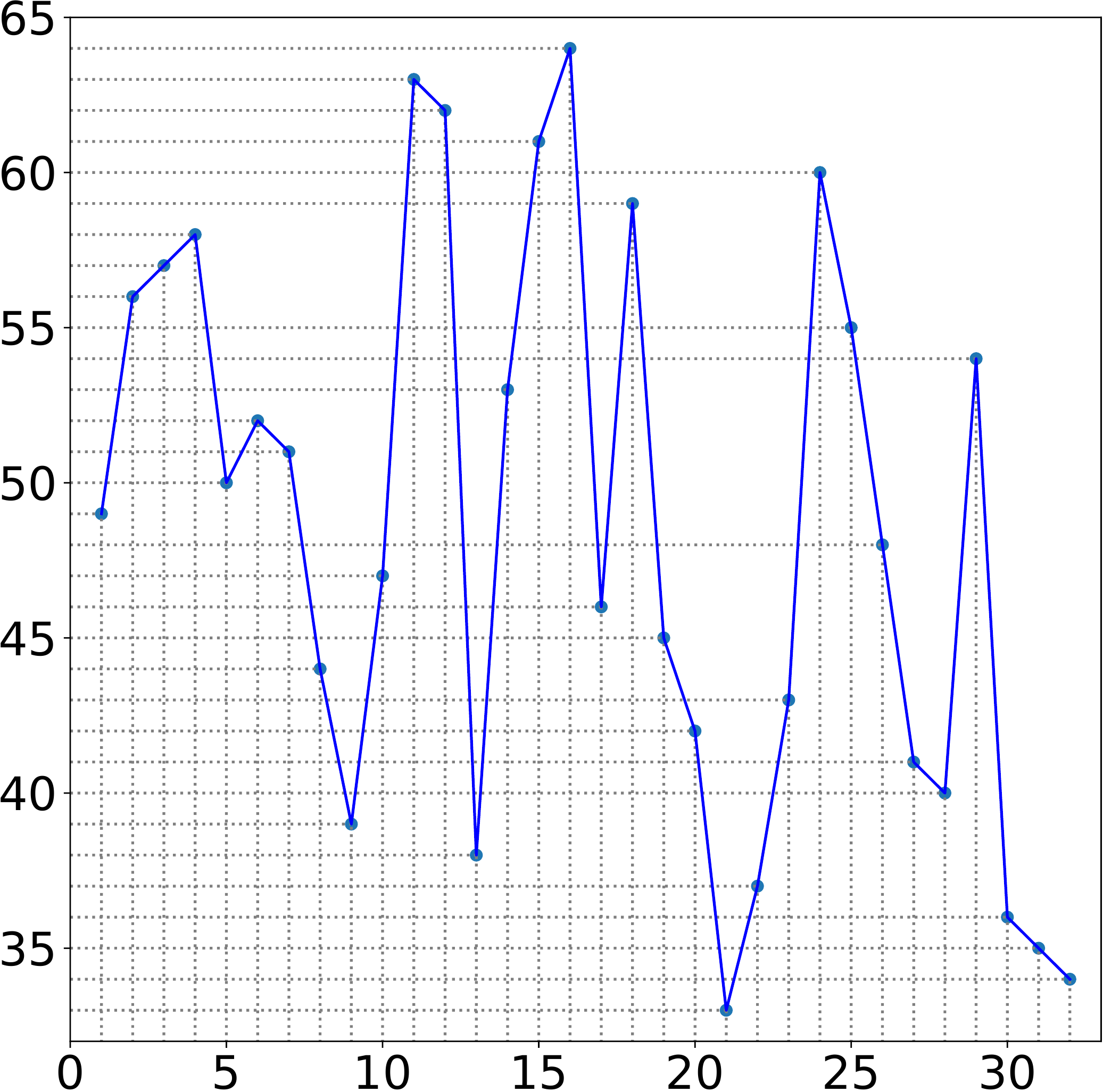}
\caption{CycleGAN Mapping}
\label{fig:sythetic-b}
\end{subfigure}
\begin{subfigure}[t]{0.33333\textwidth}
\centering
\includegraphics[width=0.5\textwidth]{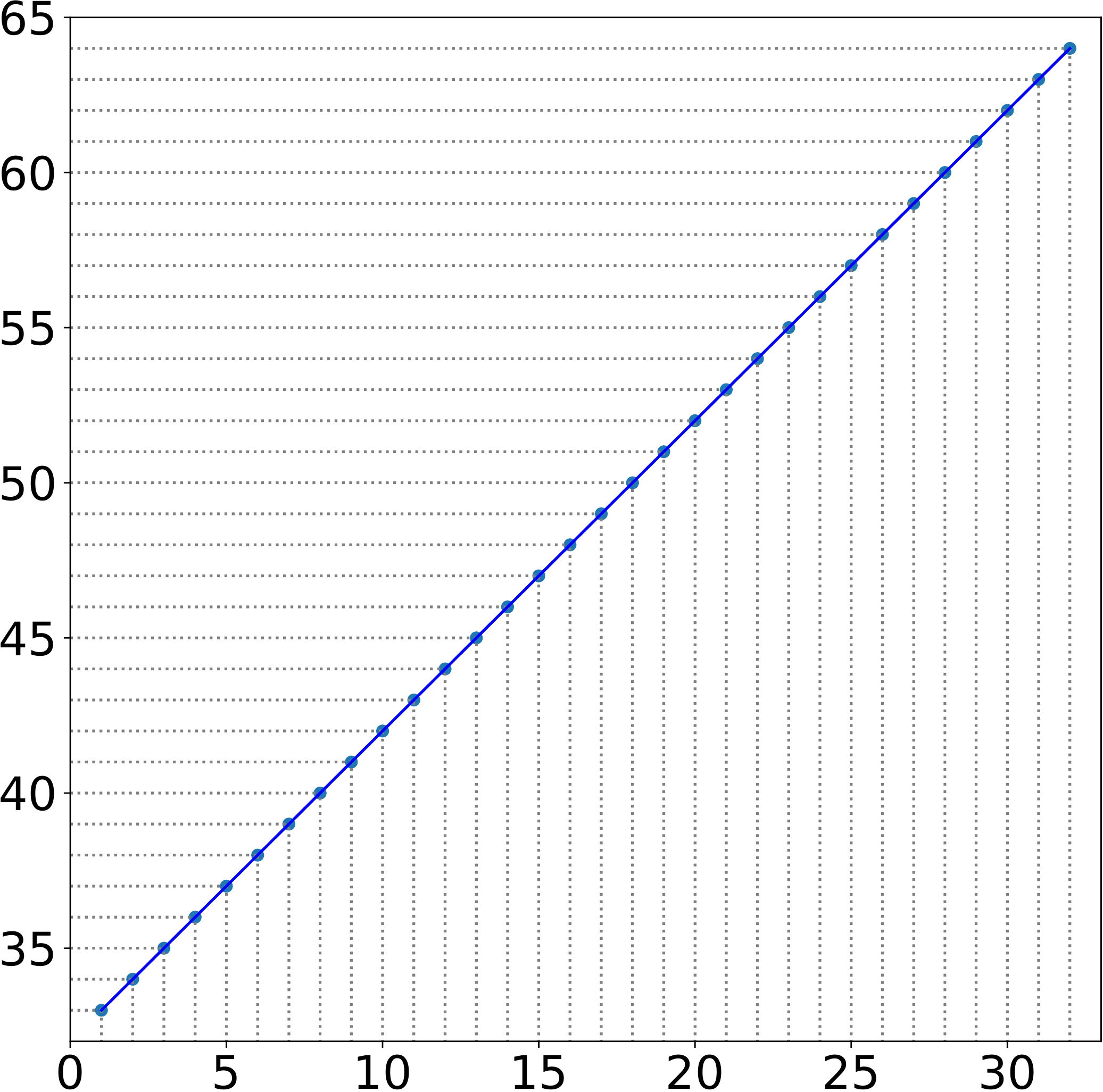}
\caption{Optimal Transport Mapping}
\label{fig:sythetic-c}
\end{subfigure}
\caption{Synthetic experiments: CycleGAN learns a one-to-one mapping between  datasets A and B, however, the learned mapping is out-of-order. Defining the cost function to be the squared Euclidean distance of the locations of their vertical lines, the optimal transport is capable of mapping the images in dataset A to B in sequence. This illustrates the randomness of the one-to-one mapping established via CycleGAN and at the same time show the ability of optimal transport to build a desired mapping, given task-specific cost function. The x- and y-axis ticks in sub-figure (b) and (c) indicate the images with the specified locations of the vertical line in domain A and B respectively.}
\label{fig:sythetic}
\end{figure*}

For verification, we conducted experiment across two synthetic datasets A and B, each consists of 32 images in the resolution of 64x64, with each image contains one vertical line at a different position as showed in Figure~(\ref{fig:sythetic-a}). The resulting mapping learned with CycleGAN is showed in Figure~(\ref{fig:sythetic-b}). As we can see, images with the vertical line in different positions in A is mapped to images in B without any order. Furthermore, this one-to-one mapping changes, given different initializations and hyper-parameters. 

\subsection{Guiding CycleGAN with Optimal Transport}

As discussed above, the one-to-one mapping learned by CycleGAN can be random in the large feasible solution space. However, in many practical applications, we would expect certain feature getting matched in the learned mapping. For example, when the two domains are different languages, one may expect the semantic information of characters keeps unchanged after translation. Without any additional control, the one-to-one mapping function learned by CycleGAN, in theory, will fail to achieve this with a very high probability (approaching one as the number of term increasing).

Here we propose to make use of the controllability in optimal transport to endow CycleGAN with the ability of learning one-to-one mapping with desired properties.

%The optimal transportation problem is to find a transport map
%Optimal transport \cite{oldandnew} finds a mapping between two distributions that transports the source distribution to the target distribution with the least transportation cost, given the cost of moving mass from any two points

%number and nouns, which means the ordering on certain properties may not be maintained. Therefore when there is need for learning a mapping function to restore the

\subsubsection{Optimal Transport (OT)}
According to Kantorovich formulation~\cite{kant}, the typical optimal transport problem can be defined as finding a mapping function $\pi$ between two distribution $\mu$ and $\nu$, which is optimal with respect to cost function $c(x,y)$, and it can be formulated as follows:
\begin{equation}
\begin{aligned}
\label{eq:OT}
\pi^* &= \mathop{\mathrm{argmin}}_{\pi \in \Pi(\mu, \nu)} \int_{\mu \times \nu} c(u,v)\pi(u,v)dudv,
\end{aligned}
\end{equation}
where $\Pi(\mu,\nu)$ denotes the collection of all probability measures on $U \times V$  with marginals $\mu$ on $U$ and $\nu$ on V, as in the primal form of Wasserstein distance. In fact, Wasserstein distance is a special form of optimal transport with the cost function $c(u, v)$ required to be a distance (a proper metric), while in optimal transport, $c(u, v)$ can be any cost function. Another difference is that, as an adversarial objective, Wasserstein distance is conducted between the distribution formed by $G_{uv}(u)$ and the target distribution $\nu$, while the optimal transport is conducted between the source distribution $\mu$ and the target distribution $\nu$. And here we will focus more on the optimal transport plan, instead of the optimal cost.

\subsubsection{Reflecting the Desired Properties with OT}

Given two distribution $\mu$ and $\nu$, CycleGAN builds a one-to-one mapping $\pi(u, v)$ between $\mu$ and $\nu$. As we discussed previously, the one-to-one mapping might be a random one in the feasible solution space. However, in the specific tasks, people actually have an expectation on the outcome of the learned one-to-one mapping, \emph{e.g.} pixel-level distance or average hue difference was expected to be low, the outline or semantic meaning was expected to be unchanged and so on.

One way to model the expectation is to define a task-specific cost function $c(x, y)$ and then the satisfaction degree of the expectation, if it is defined to be the averaged satisfaction degree of all pairing, can be modeled as the transport cost $\int_{\mu \times \nu} c(u,v)\pi(u,v)dudv$, Eq.~(\ref{eq:OT}). It follows that, given a task-specific cost function $c(x, y)$, in terms of the optimal transport, the best mapping is the $\pi^*$. 

We thus propose to solve the optimal transport problem under the task-specific cost function and use the optimal transport plan $\pi^*$ as a reference to build the one-to-one mapping in CycleGAN.

%By designing the cost function $c(x,y)$ to measure the difference on the specific property, the optimal transport plan that minimizing the overall cost of mapping distribution $u$ to distribution $v$ would reflect how should we build the mapping between $u$ and $v$. 

%under a special designed cost function, .

%\emph{s.t.} \pi \in \mathcal{P} &= \left \{ \gamma \leq 0, \int_{\Omega_t}\gamma(x,y)dy = \mu_s, \int_{\Omega_s}\gamma(x,y)dx = \mu_t \right \} \\

%Note that Eq.~\ref{eq:OT} is a Linear Program and always has a solution. Note that the mapping function found by optimal transport holds certain ordering on with respect to the cost function $c(x,y)$.  which means the optimal mapping function holds the same ordering on  the corresponding properties. 
% By careful design of  which mean optimal transport can be used as a reference towards the 

\subsubsection{Optimal Transport Plan as Reference}

Given an arbitrary cost function, the optimal transport plan is usually a many-to-many mapping, i.e. $\pi(u|v)$ and $\pi(v|u)$ is usually not a Dirac delta distribution. Therefore, it is not feasible in cross-domain translation tasks, and some previous work~\cite{joint-coupling-mapping,large-scale} attempt to use the Barycenter instead. The Barycenter of $u$ in the source distribution $\mu$ is defined to be a sample $v_b^*(u)$ in target domain V that has the minimal transport cost to its transport targets $\pi(v|u)$:
\begin{equation}
\begin{aligned}
\label{eq:barycenter}
v_b^*(u) &= \mathop{\mathrm{argmin}}_{v_b}\int_{v \sim \pi(v|u)} c(v_b, v) dv,
\end{aligned}
\end{equation}
However, the Barycenter is not guaranteed to lie in the distribution $\nu$, which in practice behaves as blurring images.

We thus proposed that, instead of directly using the optimal transport plan or the Barycenter, we train a CycleGAN and use the Barycenter of the optimal transport plan as a reference to guide the establishment of its one-to-one mapping. Given a proper weight on this regularization, CycleGAN will be able to learn a one-to-one mapping that basically follows the optimal transport plan, while at the same time, makes each translated sample lies in the target distribution under the supervision of adversarial loss. Our algorithm can then be separated into two steps:
\begin{itemize}
\item Firstly, given two distributions and a task-specific cost function, we learn an optimal transport plan between the two distributions, and we evaluate the Barycenter $v^*_b(u)$ and $u^*_b(v)$ for each sample in the two distributions.
\item Secondly, we train a CycleGAN model using these Bary-centers as references to the two cross-domain generators. The corresponding reference loss is defined as follows:
\begin{equation}
\begin{aligned}
&L_{ref}(G_{uv}) = {\mE}_{u \sim \mu}[\| G_{uv}(u) - v_b^*(u) \|], \\
&L_{ref}(G_{vu}) = {\mE}_{v \sim \nu}[\| G_{vu}(v) - u_b^*(v) \|]. \\
\end{aligned}
\end{equation}
\end{itemize}
The full objective of our algorithms can be formulated as:
\begin{equation}
\begin{aligned}
L(G_{uv}, G_{vu}&, D_u, D_v)= L_{wgan}(G_{uv},D_v) + L_{wgan}(G_{vu},D_u) \\
&- \lambda_{gp} \cdot (L_{gp}(D_{u})+L_{gp}(D_{v})) \\
&+ \lambda_{rec} \cdot (L_{rec}(G_{uv})+L_{rec}(G_{vu})) \\
&+ \lambda_{ref} \cdot (L_{ref}(G_{uv})+L_{ref}(G_{vu})), \\
 \end{aligned}
\end{equation}
where $G_{uv}$ and $G_{vu}$ are optimized to minimize the objective, while $D_v$ and $D_u$ are optimized to maximize the objective. We will later refer to this model as OT-CycleGAN.

\subsection{Discussions}

\begin{table}
% %\vspace{5pt}
\centering
\resizebox{\columnwidth}{!}{
\begin{tabular}{c|c|c|c}
\hline 
%&CycleGAN & \shortstack{Optimal \\ Transport} & Nearest \newline Neighbor \\
& \multirow{2}{*}{CycleGAN} &  Optimal & Nearest \\ && Transport & Neighbor \\
\hline
\rule{0pt}{2.0ex}
Controlling & N & Y & Y\\
\hline
\rule{0pt}{2.0ex}
Mapping & One-to-One & Many-to-Many & N/A\\
\hline
\rule{0pt}{2.0ex}
Generalization & Y & N & N \\
\hline
\end{tabular}
}
\caption{Comparison among CycleGAN, optimal transport, and nearest neighbor. The nearest neighbor and optimal transport are capable of controlling the mapping with respect to a given metric between two samples. However, the mapping build via nearest neighbor does not form a joint distribution, i.e. may collapse to a subset, and optimal transport usually builds a many-to-many mapping, which is not adequate in cross-domain translate. And also, the optimal transport plan does not generalize to out-of-distribution samples.}
\label{table1}
\end{table}

As discussed in the previous sections, in the sense of establishing a mapping between two data distributions, CycleGAN and optimal transport both have strengths and weaknesses. %CycleGAN has the promising capability of establishing a one-to-one mapping between two data distributions with the help of adversarial loss and cycle consistency loss, but it is unable to conduct further control on the property of the learned one-to-one mapping. In contrast, optimal transport finds out the optimal mapping between two data distributions which minimizing the overall cost of mismatching of desired properties, where the degree of mismatching is defined via a task-specific cost function $c(x,y)$. However, in most case, the optimal transport is a many-to-many mapping and the Barycenters might be out-of-distribution samples.
%
%Under the circumstance of two discrete distributions (discrete datasets), the probability of each sample within each distribution is the same, so that the resulting optimal mapping is totally controlled by the specified ground distance between samples from different data distributions. So the mapping found by solving the optimal transport problem is guaranteed to be 'ordered' as it aims to find out the transport plan with the least overall transportation cost to transport the source distribution to the target distribution while the one learned by CycleGAN is usually 'unordered'. \yuxuan{this part is very confused, need to be modified.}\zhiming{totally don't understand what you would like to say.}
%
This motivates us to use the barycenters of optimal transport mapping to serve as the references of CycleGAN, so as to combine the strengths of the two models to establish a one-to-one mapping with (mostly) minimized mismatching cost over task-specific properties between two data distributions.

Another difference between CycleGAN and optimal transport is that optimal transport establishes a mapping between samples in both datasets mathematically. Under the circumstance of two discrete datasets, it cannot \textbf{generalize} to out-of-distribution samples. In contrast, CycleGAN learns the mapping function between two distributions via two neural networks and thus has the ability to generalize to out-of-distribution samples. When the two discrete datasets hold the same number of unduplicated samples, a perfect one-to-one mapping actually may also exist in optimal transport. Under such conditions, CycleGAN helps optimal transport generalized to out-of-distribution samples.

%Under the circumstance of two discrete distributions, the 
Besides optimal transport, \textbf{nearest neighbor} might also come to mind for controlling the mapping to have matched properties. 
%The nearest neighbor algorithm is also controlled by the specified ground cost and for any sample in the source distribution, it aims to find the one in the target distribution with the shortest distance.
With the nearest neighbor algorithm, every sample in the source distribution will be mapped to the nearest one in the target distribution. However, nearest neighbor is a local algorithm, and without considering the global status, the mapping established via nearest neighbor might collapse to a subset in the target domain or even a single point.
%Suppose that according to a task specified distance, the support of the source domain and the support of the target domain have no overlap. 
For example, source domain is a set of real numbers whose range is [0, 31] while the range of target domain is [32, 63] and the cost function is specified as the squared difference. In this case, the nearest neighbor would map all samples in the source domain to the `leftmost' one in the target domain \emph{i.e.} 32. In comparison, optimal transport will map the whole source domain to the whole target domain in sequence.

% \begin{figure}
% \includegraphics[width=0.9\columnwidth]{figures/sandiantu_ot.pdf}
% \caption{The mapping function learned by OT-CycleGAN}
% \label{mapping_otcyclegan}
% \end{figure}
%\subsection{Network Architecture}

We summarize the discussion among CycleGAN, optimal transport and nearest neighbor in Table~(\ref{table1}).

%In practice, it's actually very common  that the  supports of two distribution are non-overlapping, especially in high-dimension situation. As aforementioned, the nearest neighbor algorithm is unable to establish a mapping between two distributions under this circumstance.  

% So we can get that when the ranges of two domains are not the same which is very common in practice, the nearest neighbor algorithm will map one domain to the subset of the other domain. Thus, nearest neighbor algorithm is unable to establish a mapping between two domains.

\section{Experiments}

In order to demonstrate the effectiveness of our proposed algorithm for learning a one-to-one mapping between two data distributions with desired properties, we conduct several image-to-image translation experiments between different datasets, and we compare the translation results of our algorithm with CycleGAN. Details of our experimental setting are as follows. 

\subsection{Network Architecture}
% In image-to-image translation, there are two kinds of features in both domain. One is the kind of features exists in both domains called background features and the other one is the kind of features exits in either domain called foreground features. 
In our experiments, we adopted the architecture of auto-encoder~\cite{autoencoder} in both of our generators. %The encoder part of each generator learns to extract the compact representation of the source samples while the decoder part learns to render it to samples in the target domain, achieving the task of image-to-image translation. 
The encoder is composed of a set of stride-2 convolution layers with a 4x4 filter, while the decoder is composed of several stride-2 deconvolution layers with 4x4 filter. Each convolution layer in the encoder or deconvolution layer in the decoder is followed by a normalization layer except the first and the last one. %The activation layers in encoder is LeakyReLU layer, while in decoder, Relu is utilized except the last one is Tanh instead. 
We use WGAN-GP loss instead of the original GAN loss in our experiments. The architecture of discriminator~(critic) is designed to be the similar as the decoder, except that we eliminate all normalization layers. % and also the last activation layer. 

\subsection{Optimization Details}
We use network simplex algorithm \cite{network_simplex} for solving the optimal transport problem between two data distributions as linear programming. Due to the lack of computation power, we use L2 barycenter instead of accurate barycenter to obtain the barycentric mapping out of the previously-obtained optimal transport plan, which can be simplified as the weighted sum of mapped samples. We use Adam~\cite{adam} optimizer with $\beta_{1}=0.5$, $\beta_{2}=0.999$. We train our model for 3000 epochs with an initial learning rate of 0.0002 and linearly decayed it to zero. $\lambda_{gp}$ is set as 10, $\lambda_{rec}$ is set in the range of [100, 800] and $\lambda_{ref}$ is set in the range of [50, 300]. We train critic for 5 steps and generator for 1 step in turn.

\begin{table*}[t!]
%\vspace{5pt}
\centering
\resizebox{1.8\columnwidth}{!}{
\begin{tabular}{c|c|c|c|c|c|c}
\hline 
& \multirow{2}{*}{CycleGAN} & \multicolumn{5}{c}{OT-CycleGAN} \\
\cline{3-7}
& & \multicolumn{1}{c|}{$\lambda_{ref}=50$} & \multicolumn{1}{c|}{$\lambda_{ref}=100$} & \multicolumn{1}{c|}{$\lambda_{ref}=200$} & \multicolumn{1}{c|}{$\lambda_{ref}=300$} & \multicolumn{1}{c}{$\lambda_{ref}=500$} \\
\hline
Mismatching Degree~($\times 10^4$) &1.026 & 0.5634 & 0.3393 & \bf0.2788 & 0.2865 & 0.3023\\
\hline
\end{tabular}
}
%\vspace{-0pt}
\caption{Comparison between CycleGAN and OT-CycleGAN in terms of mismatching degree.}
\label{table2}
%\vspace{-3pt}
\end{table*}

\subsection{Experiment: Car-to-Chair}

We conduct our first experiment between a car dataset \cite{dataset_car} and a chair dataset \cite{dataset_chair}. Both datasets consist of images of 3D rendered objects with varying azimuth angles and the value of azimuth angle of each image is provided by the dataset. %, which are several fixed values within $-90^{\circ}$ and $90^{\circ}$ at $15^{\circ}$ intervals.
Figure~(\ref{fig:car2chair-b}) shows the translation results of CycleGAN between these two datasets. As we can see, as the images of car vary in azimuth angle in order, the translation results are random samples in the target domain.

\subsubsection{OT Barycenter}
\begin{figure}[thbp]
%\vspace{-3pt}
\begin{subfigure}[t]{0.5\textwidth}
\centering
\hspace{-15pt}
\includegraphics[width=0.97\textwidth]{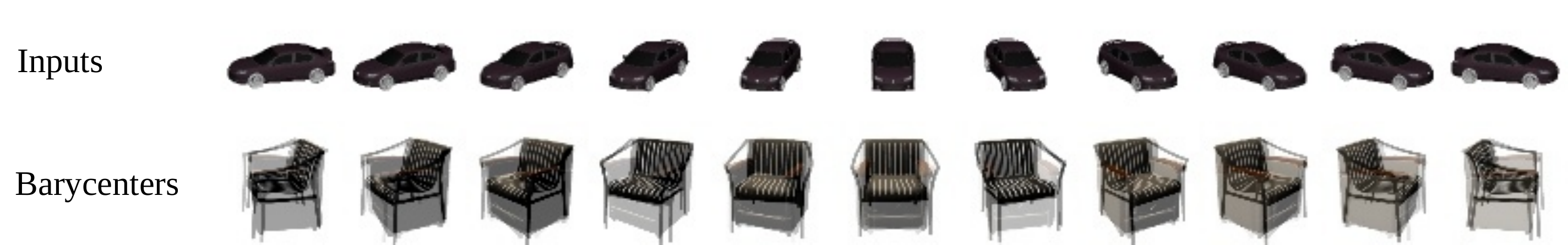}
\caption{Barycentric Mapping}
\label{fig:car2chair-a}
\end{subfigure}
\begin{subfigure}[t]{0.5\textwidth}
\centering
\hspace{-15pt}
\includegraphics[width=0.97\textwidth]{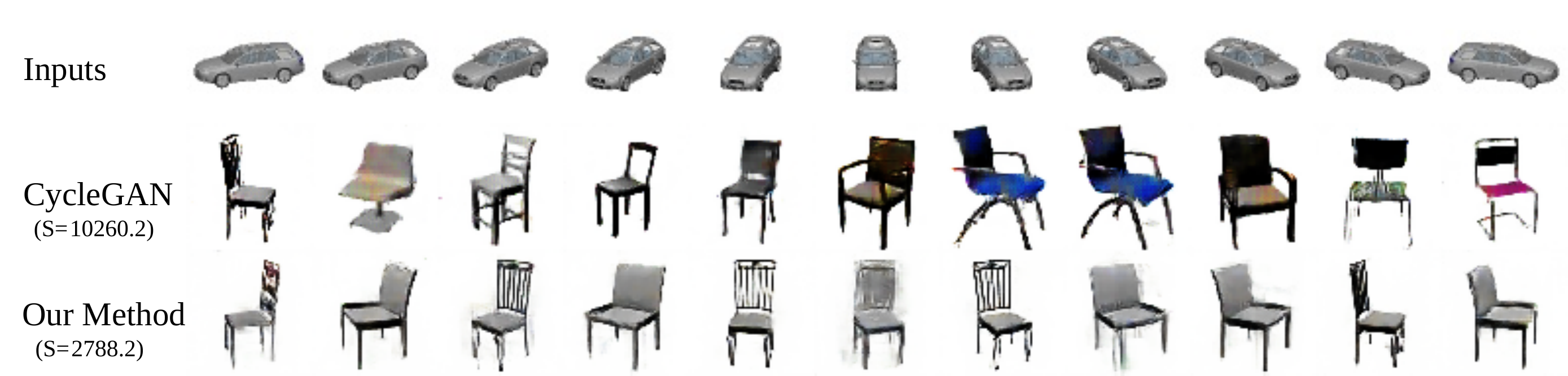}
\caption{Result Comparison}
\label{fig:car2chair-b}
\end{subfigure}
%\vspace{-3pt}
\caption{Car-to-Chair Experiments.}
\label{fig:car2chair}
%%\vspace{-5pt}
\end{figure}

By using the azimuth angle of each image provided by each dataset and specifying the cost function between each image to be the squared difference of azimuth angle, we are able to find an optimal transport plan that can transport the car distribution to the chair distribution with the least overall azimuth angle difference. Additionally, as there is more than one image at each azimuth angle, we further use the Euclidean distance between the average RGB color of each image (exclude the white background) to as subsidiary cost function, such to find an optimal transport plan that can further minimize the overall color difference. In summary, the task-specific cost function in this experiments is formulated to be:
\begin{equation}
c(x, y) = d_{angle}(x, y) + \lambda_{color} \cdot d_{color}(x,y).
\end{equation}
%where the angles is in range of $[-90^{\circ}, 90^{\circ}]$, the colors in each RGB channel is in range of [0, 255], and the weight of color distance $\lambda_{color}$ is set to be 100 in this experiment. 
The samples of resulting barycentric mapping are illustrated in Figure~(\ref{fig:car2chair-a}). %\zhiming{angle range, color range}

\subsubsection{OT-CycleGAN Result}
Figure~(\ref{fig:car2chair-b}) shows translation results of our algorithm. The resulting mapping of OT-CycleGAN successfully matches the azimuth angles and colors of the generator's input and output. % in the learned one-to-one mapping between car dataset and chair dataset with less overall azimuth and color difference. 
We also evaluate the \textbf{mismatching degree} $S=\int_{\mu \times \nu} c(u,v)\pi(u,v)dudv$ for each method. As listed in Table~(\ref{table2}), OT-CycleGAN achieves a much lower mismatching degree.

\subsubsection{Azimuth-Angle Mapping Analysis}

\begin{figure}[hbtp] 
  \begin{subfigure}[b]{0.5\linewidth}
    \centering
    \includegraphics[width=0.55\linewidth]{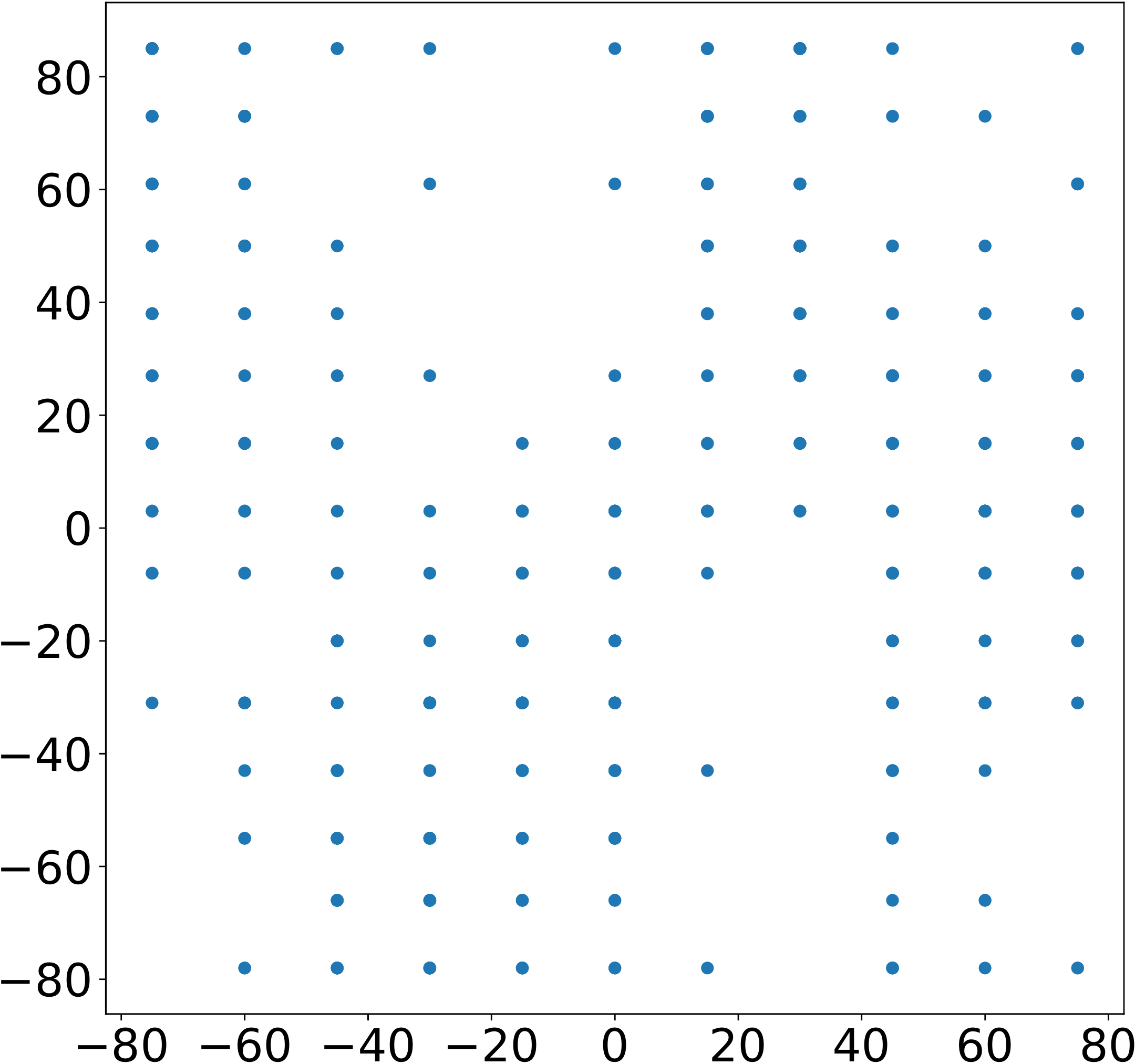} 
    \caption{Car-to-Chair: CycleGAN} 
    \label{fig7:a} 
  \end{subfigure}%% 
  \begin{subfigure}[b]{0.5\linewidth}
    \centering
    \includegraphics[width=0.55\linewidth]{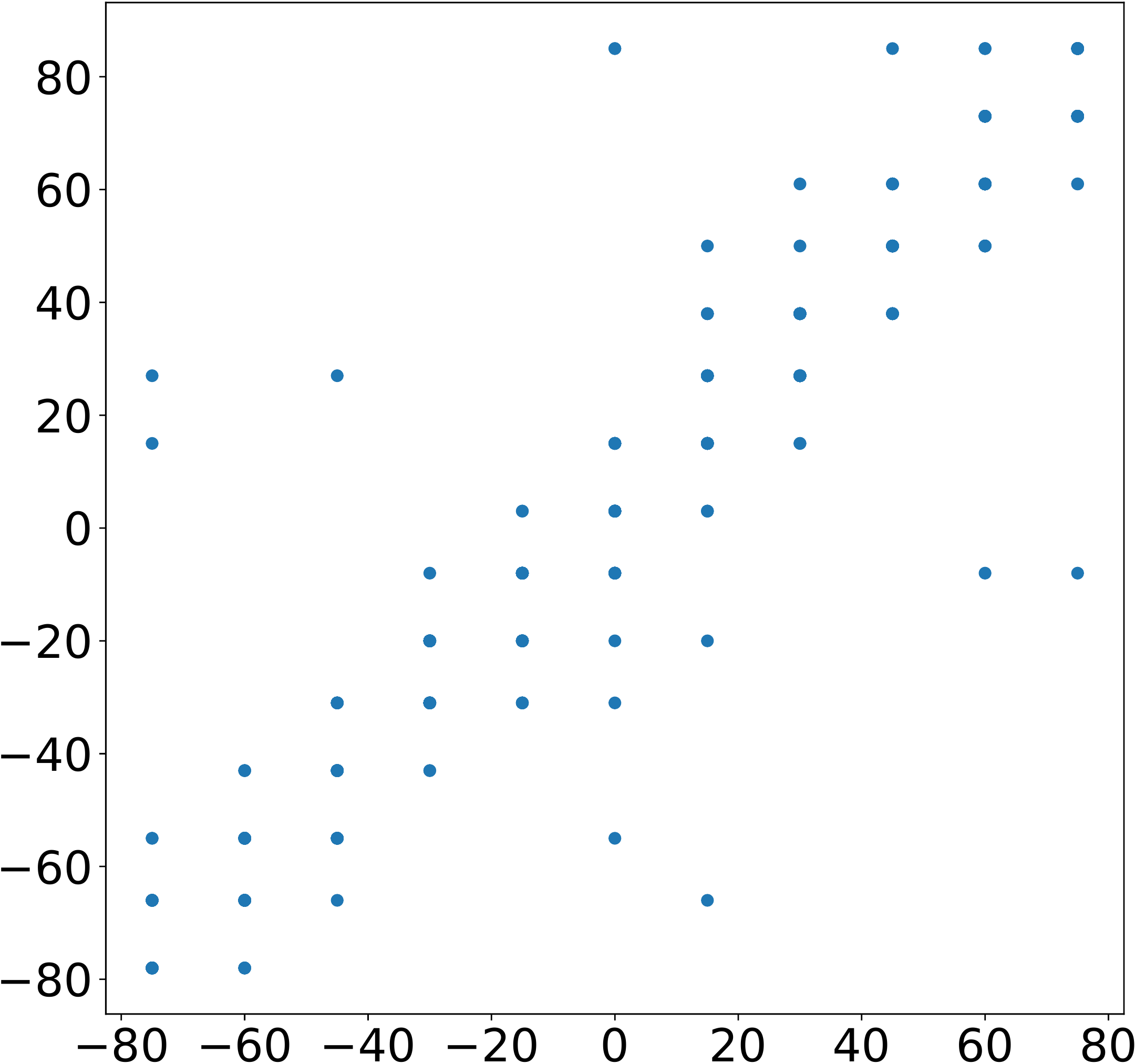} 
    \caption{Car-to-Chair: OT-CycleGAN} 
    \label{fig7:b} 
  \end{subfigure} 
  \begin{subfigure}[b]{0.5\linewidth}
    \centering
    %\vspace{5pt}
    \includegraphics[width=0.55\linewidth]{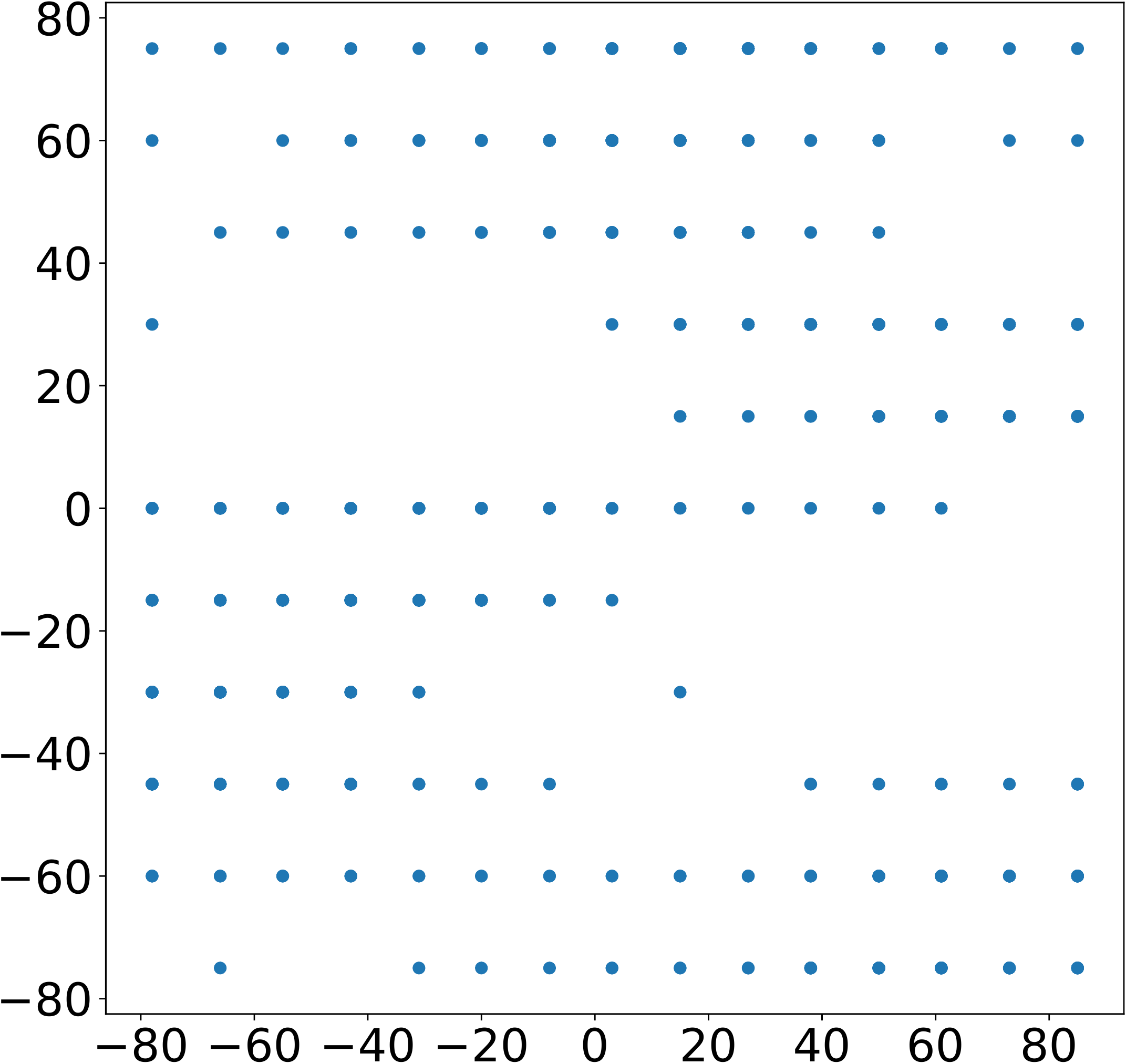} 
    \caption{Chair-to-Car: CycleGAN} 
    \label{fig7:c} 
  \end{subfigure}%%
  \begin{subfigure}[b]{0.5\linewidth}
    \centering
    %\vspace{5pt}
    \includegraphics[width=0.55\linewidth]{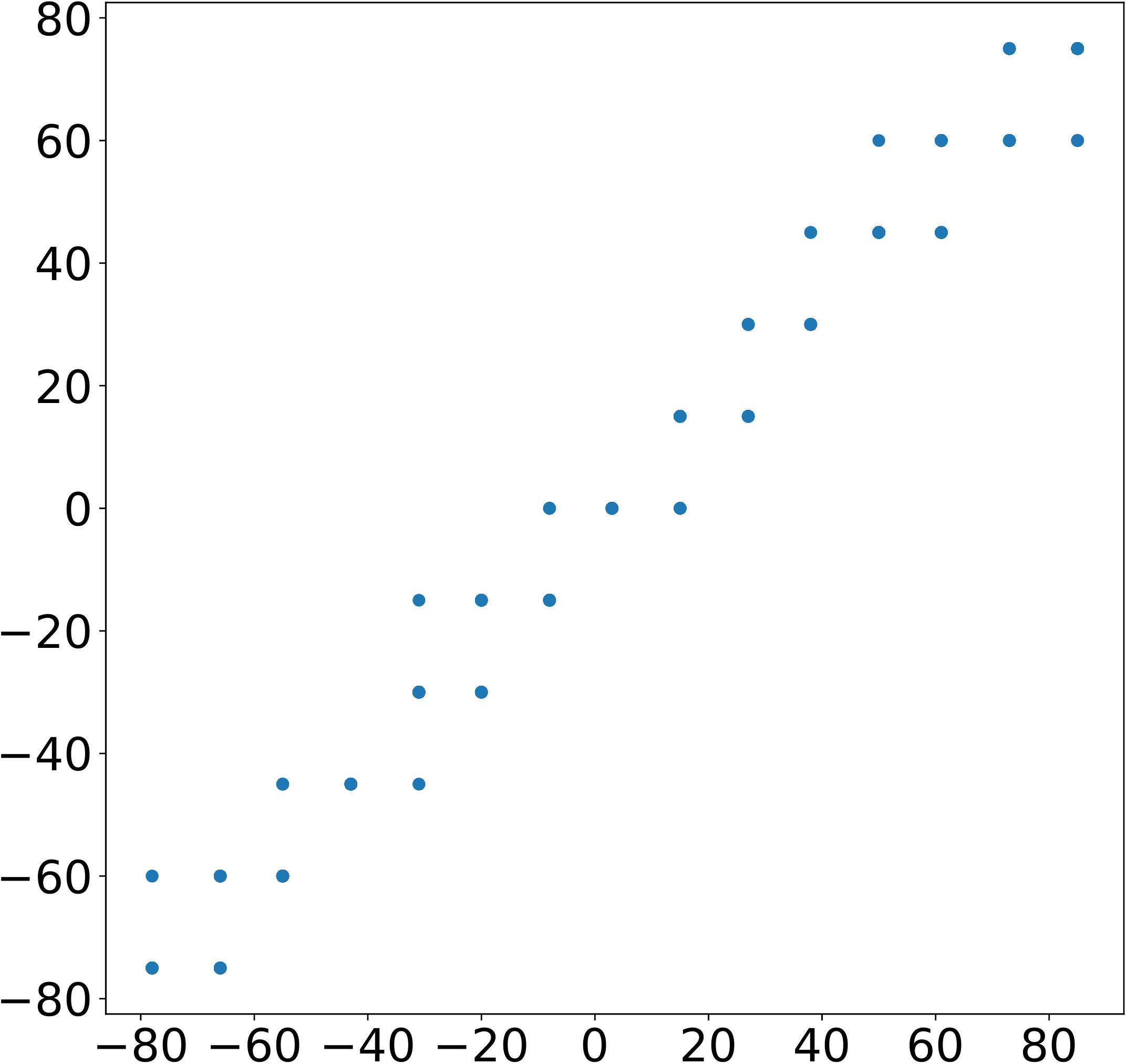} 
    \caption{Chair-to-Car: OT-CycleGAN} 
    \label{fig7:d} 
  \end{subfigure} 
  %\vspace{-13pt}
  \caption{Azimuth angle mapping of Car-to-Chair.}
  \label{fig7}
  %\vspace{-5pt}
\end{figure}

We plot the overall azimuth-angle mapping to provide a global comparison between CycleGAN and OT-CycleGAN. As we can see in Figure~(\ref{fig7}), the resulting azimuth-angle mapping with CycleGAN is fairly random, while the OT-CycleGAN mostly matches the azimuth-angle of input and output. The azimuth-angle of translated image is obtained via finding its nearest neighbor in the training set. It worth mentioning that here we ignored the color attribute, therefore, the result is a superposition over images of different colors.

\subsection{Experiment: Shoes-to-Handbags}

\begin{figure}[thbp]
\begin{subfigure}[t]{0.5\textwidth}
\centering
\hspace{-15pt}
\includegraphics[width=0.97\textwidth]{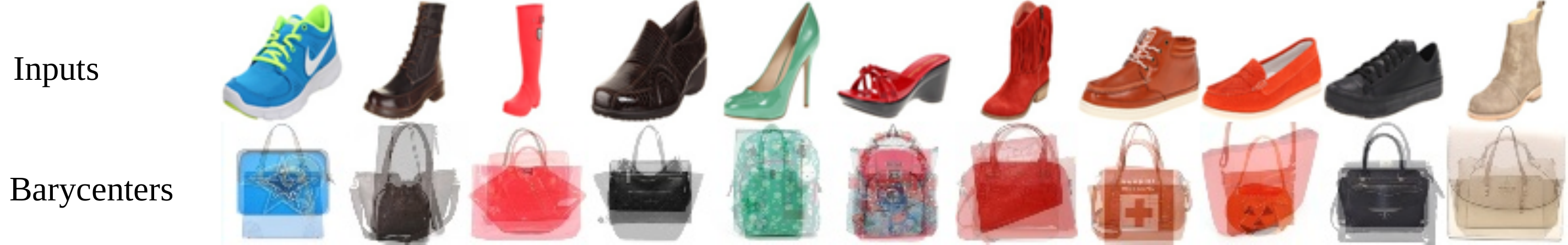}
%%\vspace{-2pt}
\caption{Barycentric Mapping}
\label{fig:shoes2handbags-a}
\end{subfigure}
\begin{subfigure}[t]{0.5\textwidth}
%\vspace{5pt}
\centering
\hspace{-15pt}
\includegraphics[width=0.97\textwidth]{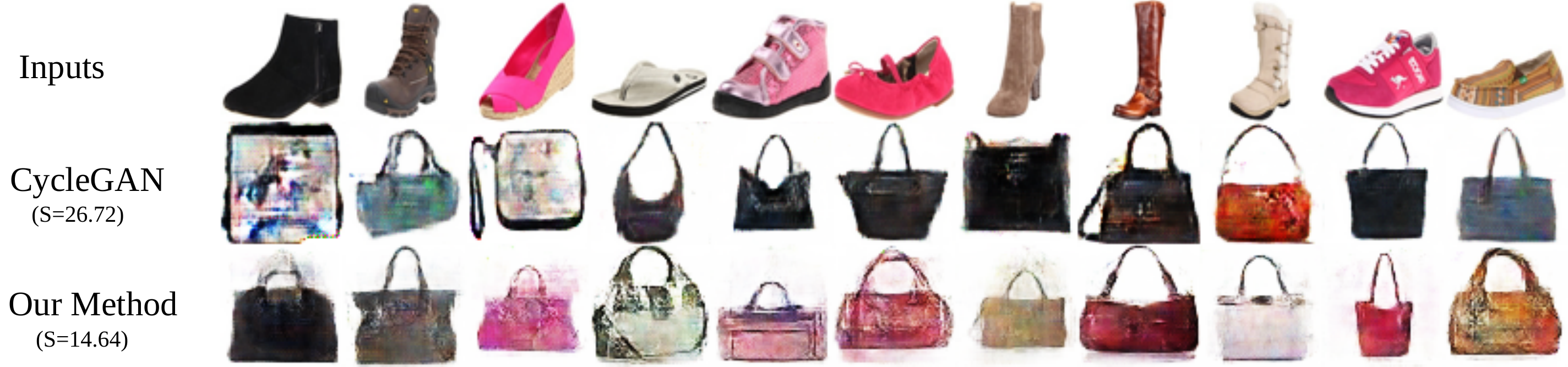}
\caption{Result Comparison}
\label{fig:shoes2handbags-b}
\end{subfigure}
%\vspace{-3pt}
\caption{Shoes-to-Handbags experiments.}
\label{fig:shoes2handbags}
%\vspace{-2pt}
\end{figure}

In this experiment, we performed image-to-image translation between a shoes dataset \cite{dataset_shoes} and a handbags dataset \cite{dataset_handbag}. %The shoes dataset consists of images of shoes collected from Zappos.com and the handbags datasets consists of images of handbags downloaded from Amazon. 
Figure~(\ref{fig:shoes2handbags-b}) shows the translation results of CycleGAN between these two datasets. As we can see, the translation results are of an obvious color difference from the source samples.

\subsubsection{OT Barycenter}

In this experiment, we would like to establish a one-to-one mapping that matches the color of the handbags with the color of shoes. As the color in each image of these two datasets is much more complex than the previously used car and chair datasets, it would be inaccurate to use the average color to represent the color information of each image, we thus adopted a color histogram to represent color information of each image. %We first converted the image into Lab color space and calculated the histogram in this color space by discrete the 3-d space by a bin size of 12.5. Then we filtered out bins with weight less than 0.1\%.
%The resulting histograms for images are of various length so  \zhiming{??various length}
We use the Wasserstein distance $W(\mu,\nu)$ between two histograms $\mu$ and $\nu$ as the cost function, with $d(u, v)$ being the Euclidean distance of two color bins $u$ and $v$ in Lab color space. %We use Euclidean distance between the color of two bins as the Lab color space is designed to be perceptually uniform with respect to human color vision so that short Euclidean distances correlate strongly with human color discrimination performance. %The average length of histogram is about 40 so that the EMD between two color histograms can be efficiently solved. 
%\zhiming{add Equation}
Samples of the resulting barycentric mapping are showed in Figure~(\ref{fig:shoes2handbags-a}). 

\subsubsection{OT-CycleGAN Result}

Figure~(\ref{fig:shoes2handbags-b}) illustrates the mapping function learned by our method~(OT-CycleGAN). Compared with the original CycleGAN, the mapping established by our algorithm is significantly better, in terms of whether the color distributions match each other, in both visual and quantitative metric $S$.

%\zhiming{Evaluate the degree of mismatching, nearest neighbor. i.e. the transport cost, add a table showing this}

%\zhiming{Show difference on barycenter, (l1-approximated barycenter,) l2-approximated barycenter for shoe and handbag.}

\subsection{Reference Weight} 

\begin{figure}[ht]
\centering
\hspace{-10pt}
\includegraphics[width=1.0\columnwidth]{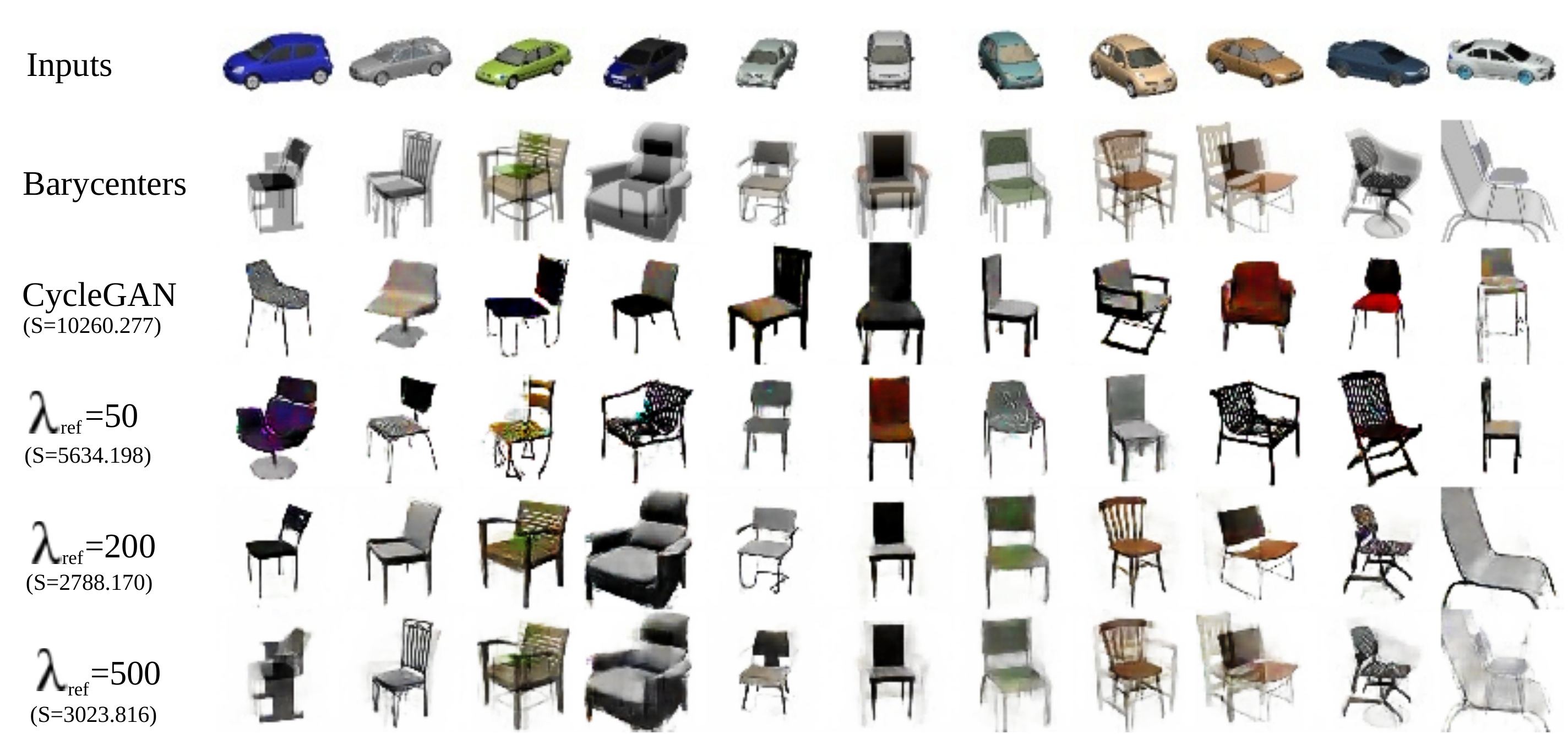}
\hspace{2pt}
\caption{Tuning the parameter $\lambda_{ref}$.}
\label{shoes2hb}
%\vspace{-8pt}
\end{figure}

One important parameter in OT-CycleGAN is $\lambda_{ref}$, i.e. the weight of OT reference loss to CycleGAN. Ideally, if $\lambda_{ref}$ is extremely large, the resulting mapping will be identical to the barycentric mapping of OT, while if the $\lambda_{ref}$ is extremely small, the reference loss will not take effect and the result will be similar to CycleGAN, which is evidenced in Figure~\ref{shoes2hb}. 
More results are summarized in Table~(\ref{table2}), and we can see there exists a pretty large range of $\lambda_{ref}$ where OT-CycleGAN is able to learn a satisfactory mapping.

% illustrates how $\lambda_{ref}$ influence the final result. As we can see, with a relatively low $\lambda_{ref}$~(e.g. 50), the mapping turns to be the random, like CyecleGAN; and when it is too high~(e.g. 500), the translation results get blurred as the barycenter. 

\subsection{Discussion} 
U-Net architecture is mostly used in image-to-image translation tasks, though it tends to connect the pixel information between the input and output and achieved many satisfactory results, it, however, does not theoretically guarantee the relationship between source and target and thus may require extensive tuning if a special property is wanted. Our method, in contrast, can directly specify which properties to be matched.

%\zhiming{L1, L2 reconstruction; semisupervised; large scale OT; Berycenter; Other task; U-net and Auto-encoder.}

\section{Conclusion and Future Work}
We have presented OT-CycleGAN where an optimal transport mapping is used to guide the one-to-one mapping established by CycleGAN. With the proposed algorithm, one can control the learned one-to-one mapping in CycleGAN via defining a task-specific cost function that reflecting the desired mapping properties.

Specifically, we demonstrate that there is no controllability on the properties of the learned one-to-one mapping in CycleGAN, and optimal transport can provide a mapping that minimizing the overall cost of mismatching of expected properties, given a task-specific cost function. Since the optimal transport mapping is usually not one-to-one, we propose to use the Barycenters of learned mapping as references to guide the training of CycleGAN to form a one-to-one mapping with desired mapping properties. 

Experiments conducted on several benchmark datasets have shown that the mapping function learned by vanilla CycleGAN can be quite messy and the guiding of optimal transport can significantly improve the mapping in terms of the task-specific properties.

In the main-body and experiments, we mainly focused on image-to-image translation, as it is the most successful application of CycleGAN. We hope the detailed analysis of the properties of CycleGAN and optimal transport would also benefit further investigation on cycle consistency loss and unsupervised cross-domain translation. OT-CycleGAN is a general framework for establishing one-to-one mapping with desired properties and we plan to investigate more related tasks in the further.
 
\bibliography{reference}
\bibliographystyle{aaai}
\end{document}